\documentclass[letterpaper, 10 pt, conference]{ieeeconf}  

\IEEEoverridecommandlockouts                \usepackage{soul} 
\usepackage{xcolor}

\title{\LARGE \bf
Parallel Optimization with Hard Safety Constraints for Cooperative Planning of Connected Autonomous Vehicles}

\author{Zhenmin Huang, Haichao Liu, Shaojie Shen, and Jun Ma
\thanks{This work was supported in part by the National Natural Science Foundation of China under Grant 62303390; and in part by the Project of Hetao Shenzhen-Hong Kong Science and Technology Innovation Cooperation Zone under Grant HZQB-KCZYB-2020083. \textit{(Corresponding Author: Jun Ma.)}}
\thanks{Zhenmin Huang is with the Robotics and Autonomous Systems Thrust, The Hong Kong University of Science and Technology (Guangzhou), Guangzhou, China, and also with the Division of Emerging Interdisciplinary Areas, The Hong Kong University of Science and Technology, Hong Kong SAR, China (email: zhuangdf@connect.ust.hk).}
\thanks{Haichao Liu is with the Robotics and Autonomous Systems Thrust, The Hong Kong University of Science and Technology (Guangzhou), Guangzhou, China (e-mail: hliu369@connect.hkust-gz.edu.cn).}
\thanks{Shaojie Shen is with the Department of Electronic and Computer Engineering, The Hong Kong University of Science and Technology, Hong Kong SAR, China (email: eeshaojie@ust.hk).}
\thanks{Jun Ma is with the Robotics and Autonomous Systems Thrust, The Hong Kong University of Science and Technology (Guangzhou), Guangzhou, China, also with the Division of Emerging Interdisciplinary Areas, The Hong Kong University of Science and Technology, Hong Kong SAR, China, and also with the HKUST Shenzhen-Hong Kong Collaborative Innovation Research Institute, Futian, Shenzhen, China (e-mail: jun.ma@ust.hk).}}

\usepackage{amsmath}
\usepackage{algorithm}
\usepackage{algpseudocode}
\usepackage{graphicx}
\usepackage{subfigure}
\usepackage{multirow}
\begin{document}

\maketitle
\thispagestyle{empty}
\pagestyle{empty}
\newtheorem{remark}{Remark}
\newtheorem{definition}{Definition}

\begin{abstract}
The development of connected autonomous vehicles (CAVs) facilitates the enhancement of traffic efficiency in complicated scenarios. 
Difficulties remain unsolved in developing an effective and efficient coordination strategy for CAVs.
In this paper, we formulate the cooperative autonomous driving task of CAVs as an optimal control problem with safety conditions enforced as hard constraints, and propose a computationally-efficient parallel optimization framework to generate strategies for CAVs with the travel efficiency improved and the hard safety constraints satisfied.
Specifically, all constraints involved are addressed appropriately with convex approximation, such that the convexity property of the reformulated optimization problem is exhibited.
Then, a parallel optimization algorithm is presented to solve the reformulated optimization problem, with an embodied iterative nearest neighbor search strategy to determine the optimal passing sequence. 
It is noteworthy that the travel efficiency is enhanced and the computation burden is considerably alleviated with the proposed innovation development. 
We also examine the proposed method in CARLA simulator and perform thorough comparisons to demonstrate the effectiveness and efficiency of the proposed approach.
\end{abstract}

\section{Introduction}

Recent developments in artificial intelligence and information technologies prompt the emergence of connected autonomous vehicles (CAVs), which possess the ability to exchange information among vehicles through real-time communication technology such as 5G network and Wi-Fi~\cite{deng2019cooperative}. In this sense, cooperative autonomous driving is enabled, which points to the alleviation of traffic congestions induced by non-cooperative driving manners, and also the decrease of traffic accidents~\cite{liu2017distributed}. Although the physical basis is well-prepared, a series of problems remain largely unsolved in developing a mature coordination scheme that is suitable for cooperative autonomous driving in complicated urban traffic scenarios. Rule-based, optimization-based, and learning-based methods are intensively researched to produce various coordination strategies for different urban traffic scenarios, such as on-ramps merging~\cite{hang2021cooperative,rios2016automated,ding2019rule,xu2019grouping} and intersection crossing~\cite{mirheli2019consensus,fayazi2018mixed}. However, existing methods either merely focus on simple scenarios or only consider simple interactions between a limited number of CAVs, and therefore are not readily applicable to complicated urban traffic scenarios with dense traffic flow. Besides, a common problem in existing methods is the low computation efficiency. As the number of CAVs involved in the scenario increases, the scale of the problem continues to grow, which leads to a rapid surge in the overall computation time that violates the real-time requirements for planning and control of CAVs. As a result, current methods generally fail to apply to scenarios containing a large number of CAVs.

In this paper, we focus on the development of an optimization framework to solve the cooperative autonomous driving problem for all-directional traffic flows in urban traffic scenarios. To provide hard safety guarantees, a linearization method is introduced to properly handle the complicated nonlinear collision avoidance constraints and road boundary constraints. To improve travel efficiency, a novel coordination strategy based on the iterative nearest neighbor search is proposed to determine the optimal passing sequence. In conjunction with a parallel optimization method based on dual consensus ADMM, the computational burden is greatly alleviated and therefore real-time performance is achieved. In terms of experimental validation, we compare our method with several baselines to demonstrate both the improved travel efficiency and the superiority in computational efficiency.

\section{Related Works}

Various methods are intensively investigated in developing the strategy for cooperative driving in urban traffic scenarios. Owing to the rapid development of data science and artificial intelligence, learning-based methods are currently progressing by leaps and bounds, which leverage real-world data to boost performance. It is declared that learning-based methods are effective to address difficult traffic scenarios~\cite{chen2019model}. In~\cite{guan2020centralized}, a centralized coordination scheme using reinforcement learning is proposed, which allows for the cooperative crossing of CAVs at an unsignalized intersection. In~\cite{toghi2021cooperative}, multi-agent reinforcement learning is adopted to enable sympathy between CAVs and human drivers for efficient highway merging. Though with great potential, learning-based methods generally require a large amount of real-world data and they could suffer from poor interpretability, which hinders their wide applications in real-world scenarios.

In contrast to learning-based methods, rule-based and optimization-based methods are more mature and are capable of generating predictable and explainable results with limited computational resources. Particularly, rule-based methods mainly focus on the determination of passing orders of CAVs using handcrafted rules and criteria, followed by a low-level planner to generate trajectories for all CAVs. In~\cite{liu2017distributed}, a communication-enabled distributed mechanism is proposed to resolve conflicts for CAVs in intersections. Each vehicle solves a conflict graph locally to compute the desired time slot for passing, followed by a motion planner to determine the speed profile. In~\cite{pan2022convex}, a centralized convex optimal control framework is proposed with a hierarchical coordination scheme to enable unsignalized intersection crossing for CAVs. In~\cite{xu2019tree}, a method based on Monte Carlo tree search (MCTS) and heuristic rules is proposed to find nearly global optimal orders for CAVs to pass unsignalized intersections. Although rule-based methods are straightforward, it is generally hard to cover various corner cases with man-made rules, resulting in difficulties for rule-based methods to generalize to complicated scenarios.

Meanwhile, optimization-based methods typically formulate the cooperative driving as a constrained optimization problem and adopt a unified optimization framework to plan directly for all CAVs, thus being more concise and generalizable~\cite{kessler2019cooperative,esterle2020optimal}. In~\cite{burger2018cooperative}, an approach based on mixed-integer quadratic programming (MIQP) is presented to obtain globally optimal solutions for cooperative driving of CAVs in general on-road scenarios. In~\cite{mirheli2019consensus}, the cooperative trajectory planning problem is formulated as a mixed integer nonlinear program (MINLP) to determine conflict-free trajectories. 
However, these methods are only applicable to a small number of vehicles, as the computation time soon becomes unaffordable when the number of vehicles becomes large. To improve computation efficiency, distributed and parallel optimization methods are strongly desired. The alternating direction method of multipliers (ADMM)~\cite{boyd2011distributed} provides a general distributed framework to accelerate the optimization process, which triggers its recent applications on cooperative autonomous driving~\cite{rey2018fully}. In~\cite{wang2018parallel,zhang2021semi}, ADMM is used to split independent vehicle constraints from coupled collision-avoidance constraints between vehicles such that they can be dealt with separately, yielding a partially parallel optimization scheme.
In~\cite{saravanos2022distributed}, a fully decentralized and parallel optimization framework is proposed to solve the cooperative trajectory planning problem for general types of robots. However, its computational efficiency is still far from being satisfying, as minutes are taken for solving the problem involving only several robots. Based on dual consensus ADMM~\cite{banjac2019decentralized}, a parallel optimization framework is proposed in~\cite{huang2023decentralized}, which enables cooperative trajectory planning of CAVs with high computation efficiency. However, it adopts soft penalties for collision avoidance, which results in possible collisions or over-conservative strategies if penalties are not set appropriately. 
Besides, the devised approach neglects certain pertinent constraints for driving safety such as road boundary limitations. These problems render the generated strategy potentially unsafe in real-world situations.
Also, the time profiles for all CAVs are preassigned non-cooperatively and fixed, which may cause difficulties in resolving conflicts for situations like lane merging.

\section{Problem Statement}
\subsection{Vehicle Kinematics and Physical Constraints}
For a traffic scenario involving $N$ CAVs passing simultaneously, we use $\mathcal{N}=\{1,2,...,N\}$ to denote the set containing indices of all CAVs. We assume a complete graph structure on the communication network, and all CAVs involved possess the same kinematics, which can be expressed as
\begin{equation}
\label{shortDynamics}
    x^i_{\tau+1} = f(x^i_\tau,u^i_\tau),
\end{equation}
where $x^i_\tau$ and $u^i_\tau$ are the state vector and the control input vector corresponding to vehicle $i$ at timestamp $\tau$, $\tau\in\mathcal{T}=\{0,1,...,T-1\}$, and $T$ is the length of the horizon. In particular, we consider a state vector of four state variables $x^i_\tau=(p^i_{x,\tau},p^i_{y,\tau},\theta^i_\tau,v^i_\tau)$ and an input vector constituted by two control inputs $u^i_\tau=(\delta^i_\tau,a^i_\tau)$, where $p^i_{x,\tau}$ and $p^i_{y,\tau}$ denote the $X$ and $Y$ coordinates of the midpoint of the rear axle in the global frame, $\theta^i_\tau$ denotes the heading angle, $v^i_\tau$ denotes the velocity, $\delta^i_\tau$ represents the steering angle, and $a^i_\tau$ represents the acceleration. With the above notations, we adopt the following kinematic model~\cite{tassa2014control} for vehicle $i$:
\begin{equation}
\label{dynamics}
\left\{
\begin{aligned}
p^i_{x,\tau+1} &= p^i_{x,\tau}+f_r(v^i_\tau,\delta^i_\tau)\cos(\theta^i_\tau),\\
p^i_{y,\tau+1} &= p^i_{y,\tau}+f_r(v^i_\tau,\delta^i_\tau)\sin(\theta^i_\tau),\\
\theta^i_{\tau+1} &= \theta^i_{\tau}+\arcsin\left(\tau_s v^i_\tau\sin(\delta^i_\tau)/b\right),\\
v^i_{\tau+1} &= v^i_\tau+\tau_sa^i_\tau.
\end{aligned}
\right.
\end{equation}
Note that $b$ is the vehicle wheelbase and $\tau_s$ is the time interval. The function $f_r(v,\delta)$ is defined as
\begin{equation}
f_r(v,\delta) = b+\tau_sv\cos(\delta)-\sqrt{b^2-(\tau_sv\sin(\delta))^2}.
\end{equation}

Meanwhile, the following constraints are imposed on the control inputs:
\begin{equation}
\label{inputs}
    a_\textup{min}\leq a^i_\tau\leq a_\textup{max},\ 
    \delta_\textup{min}\leq\delta^i_\tau\leq\delta_\textup{max}.
\end{equation}

\subsection{Collision Avoidance Constraints}
To ensure driving safety, constraints on minimal distances between CAVs must be imposed to avoid any possible collision. We make the assumption that all CAVs are of the same size, and approximate each vehicle with two identical circles aligned in the longitudinal direction. The front circle and the rear circle of vehicle $i$ are denoted as $FC_i$ and $RC_i$, respectively.  With such an approximation, a sufficient condition for collision avoidance is that the distance between the centers of any pair of circles on different vehicles is no less than the diameter. We denote the center positions of $FC_i$ and $RC_i$ at time $\tau$ as $p^i_{f,\tau}$ and $p^i_{r,\tau}$, respectively. The collision-free condition is thus expressed as
\begin{equation}
\label{collisions}
||p^i_{\beta,\tau}-p^j_{\gamma,\tau}||_2\geq d_\textup{safe}.
\end{equation}
Notice that $d_\textup{safe}$ is the safe distance to maintain, which is set to be equal to the common diameter of all circles.

\subsection{Road Boundary Constraints}
To ensure that all CAVs are driving within the free space, road boundary constraints must be imposed. A sufficient condition for the satisfaction of road boundary constraints is that the minimal distance from the centers of circles to the corresponding closest road boundaries should always be no less than the radius of all circles. Equivalently, we shift all road boundaries towards the internal of the free space by a bias that equals the common radius of all circles. As a result, a new set $\mathcal{C}$ is formed, which is a subset of the free space. Then the following constraints are established:
\begin{equation}
\label{boundaries}
p^i_{\beta,\tau}\in\mathcal{C}.
\end{equation}

\subsection{Overall Formulation}
For CAVs, we set the goal as the minimization of the deviation of actual trajectories from reference trajectories, as well as the control inputs. Integrating all the constraints mentioned above, we obtain the optimal control problem as below:
\begin{equation}
\label{OCP}
\begin{array}{cl}
\min\limits_{\{u^i_\tau\}}&\sum\limits_{i=1}^N\left(\sum\limits_{\tau=0}^T||x^i_\tau-x^i_{\tau,ref}||^2_{Q^i_\tau}+\sum\limits_{\tau=0}^{T-1}||u^i_\tau||^2_R\right)\\
\textup{s.t.}&(\ref{shortDynamics}),(\ref{inputs}),(\ref{collisions}),(\ref{boundaries}),\\
   &\forall \beta,\gamma\in\{f,r\},\ i,j\in\mathcal{N},\ i\neq j,\ \tau\in\mathcal{T}.
\end{array}
\end{equation}
$x^i_{\tau,ref}$ is the reference point for vehicle $i$ at timestamp $\tau$, which contains reference waypoint, reference heading angle, and reference velocity. $||\boldsymbol{\cdot}||_Q$ and $||\boldsymbol{\cdot}||_R$ denote the $L_2$-norm weighted by matrices $Q$ and $R$, where $Q$ is positive semi-definite and $R$ is positive definite.

\section{Methodology}

In this section, we propose an optimization framework for solving problem (\ref{OCP}) to obtain the optimal cooperative driving strategy. 
 With a slight abuse of notations, we define the current nominal trajectories as $\{x^i_\tau,u^i_\tau\}$, for $\tau\in\mathcal{T}$ and $i\in\mathcal{N}$. Our goal is to search for the optimal variations $\{\delta x^i_\tau, \delta u^i_\tau\}$ on the nominal trajectories that minimize the objectives while satisfying the constraints. This process needs to be iterated until trajectories for all CAVs finally converge.

\subsection{Linearization of Hard Safety Constraints}

Safety constraints involved in urban traffic scenarios include collision avoidance constraints and road boundary constraints. It should be noted that both are imposed on the center positions of $FC$ and $RC$, while the optimization variables are bound to the rear axle centers of CAVs. This gap needs to be bridged by the Jacobians in-between, namely
\begin{equation}
\label{bridge}
\delta p^i_{\beta,\tau}=J^i_{\beta,\tau}\delta x^i_\tau.
\end{equation}

The original collision avoidance constraints are obviously non-convex and nonlinear, which could lead to difficulties in finding a feasible solution. Similar to~\cite{rey2018fully}, a straightforward method to render these constraints linear is adopted. For each pair of circles between vehicle $i$ and vehicle $j$, we construct a unit vector $\vec{n}^{ij}_{\beta\gamma,\tau}$ pointing from the center of one circle to the other. Then the following inequality provides a sufficient condition for collision avoidance between the two circles:
\begin{equation}
\label{collisionAvoidanceAll}
    \vec{n}^{ij,\top}_{\beta\gamma,\tau}(J^i_{\beta,\tau}\delta x^i_{\tau}-J^j_{\gamma,\tau}\delta x^j_{\tau})+d^{ij}_{\beta\gamma,\tau}-d_\textup{safe}\geq 0.
\end{equation}

\begin{figure}[t]
\centering
\includegraphics[scale=0.50]{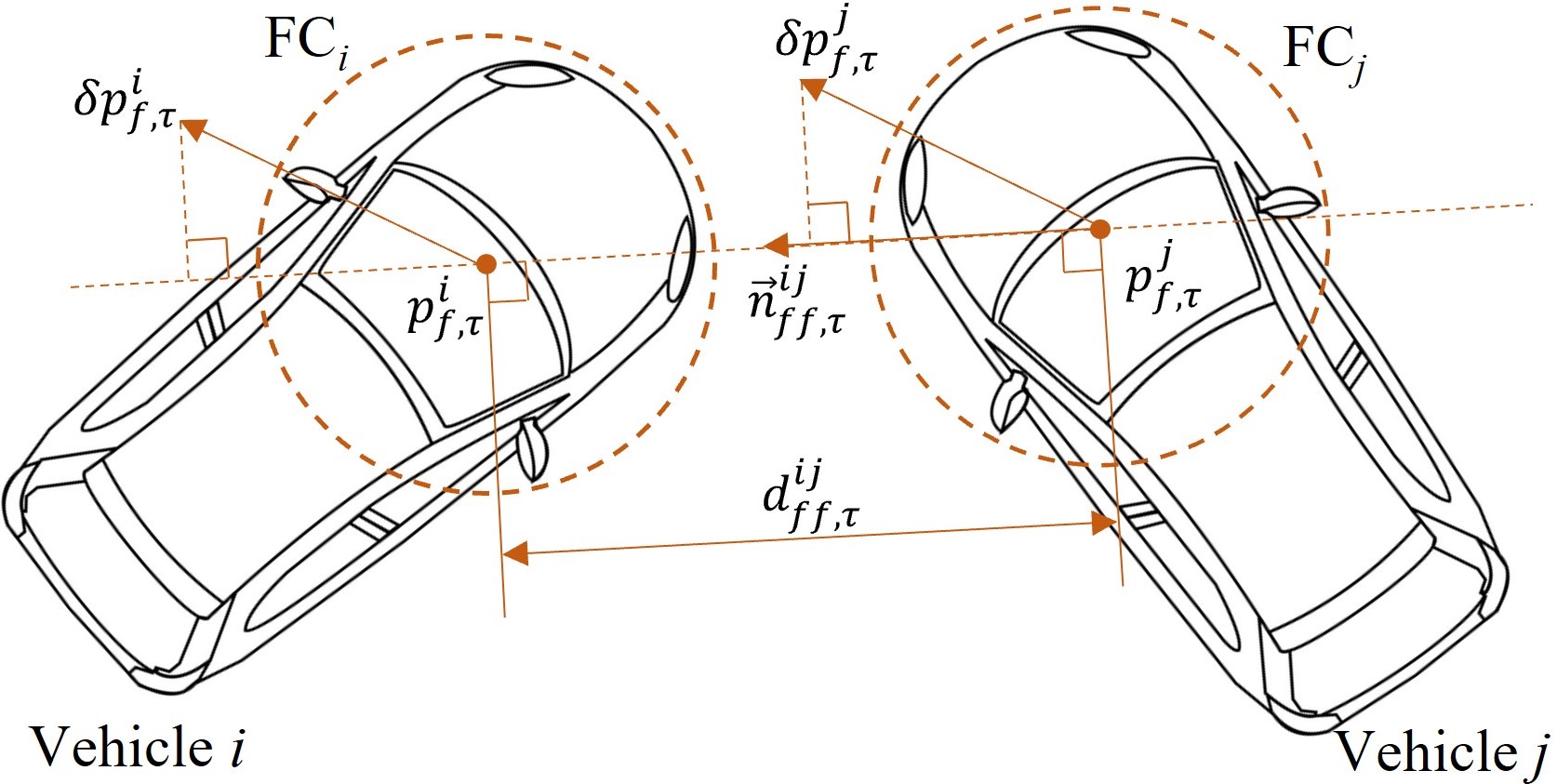}
\caption{Linearization of collision avoidance constraints.}
\label{fig:collsion_avoidance}
\end{figure}

The principle behind this inequality is that the difference between the projections of $\delta p^i_{\beta,\tau}$ and $\delta p^j_{\gamma,\tau}$ onto $\vec{n}^{ij}_{\beta\gamma,\tau}$ gives the lower bound of change in current distance $d^{ij}_{\beta\gamma,\tau}$, as is shown in Fig. \ref{fig:collsion_avoidance}. Therefore, the collision-free condition is satisfied when the sum of this difference and the current distance is greater than or equal to $d_\textup{safe}$.

To handle irregular road boundaries, we propose a linearization method that can theoretically be applied to road boundary constraints of any shape. We first extract all road boundaries in the scenario and then perform dense sampling on the boundaries to turn them into a point cloud, denoted as $\mathcal{M}$. Then, the nearest neighbor search is performed on $\mathcal{M}$ to obtain the closest boundary point corresponding to the vehicle location at each timestamp:
\begin{equation}
    p^i_{\beta b,\tau}\leftarrow NearestNeighbor(\mathcal{M},p^i_{\beta,\tau}),
\end{equation}
where $p^i_{\beta b,\tau}$ is the point closest to $p^i_{\beta,\tau}$ within $\mathcal{M}$.
We then construct the unit vector pointing from $p^i_{\beta b,\tau}$ to $p^i_{\beta,\tau}$, denoted as $\vec{n}^i_{\beta,\tau}$. With the same logic of (\ref{collisionAvoidanceAll}), the linearized road boundary constraints can be expressed as
\begin{equation}
\label{boundaryConstraints}
    2\vec{n}^{i,\top}_{\beta,\tau}J^i_{\beta,\tau}\delta x^i_\tau+2d^i_{\beta,\tau}-d_\textup{safe}\geq 0.
\end{equation}

\begin{figure}[t]
\centering
\includegraphics[scale=0.50]{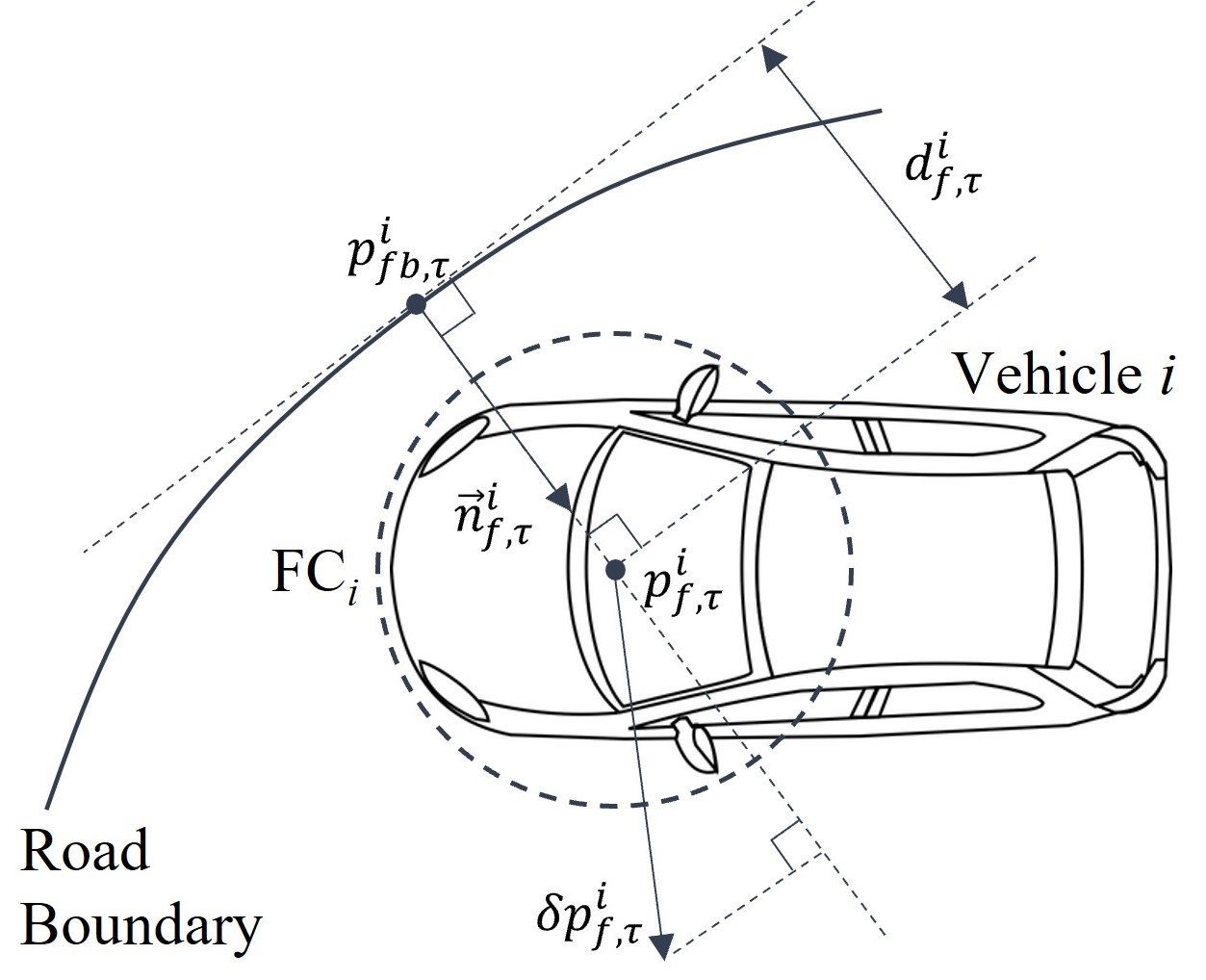}
\caption{Linearization of road boundary constraints.}
\label{fig:road_boundary}
\end{figure}

Fig. \ref{fig:road_boundary} demonstrates the linearized road boundary constraints. To obtain the closest point efficiently, we perform the nearest neighbor search using the k-d tree, which exhibits $O(logN)$ complexity for a single query to a point cloud containing $N$ points. 

\subsection{Implicit Passing Sequence Coordination}
It is a common situation in urban traffic scenarios that two vehicles try to merge into one lane simultaneously. This gives rise to the critical problem of determining the passing sequence. In this part, we propose a novel strategy that obtains the optimal passing sequence implicitly through iterative constructions of objective functions. 
We first define $\delta X^i=(\delta x^i_0,\delta u^i_0,...,\delta x^i_T)$ as the concatenated vector of all variational states and inputs corresponding to vehicle $i$. From (\ref{OCP}), the objective function corresponding to vehicle $i$ is 
\begin{equation}
\begin{aligned}
    C^i(\delta X^i) &= \sum\limits_{\tau=0}^T||\delta x^i_\tau||^2_{Q^i_\tau} + 2(x^i_\tau-x^i_{\tau,ref})^\top {Q^i_\tau}\delta x^i_\tau\\
    &+\sum\limits_{\tau=0}^{T-1}||\delta u^i_\tau||^2_R+2u^{i\top}_\tau R\delta u^i_\tau.
\end{aligned}
\end{equation}

Since the passing order of vehicle $i$ is not determined, the reference point $x^i_{\tau,ref}$ corresponding to timestamp $\tau$ is unknown. To obtain the optimal reference waypoint, we adopt the nearest neighbor search. Suppose the reference path generated by the navigation module for vehicle $i$ is $\{p^i_{ref}\}$, which is the set of all waypoints contained in the path. The current reference waypoint for time $\tau$ is given as
\begin{equation}
    p^i_{\tau,ref}\leftarrow NearestNeighbor(\{p^i_{ref}\},p^i_\tau),
\end{equation}
where $p^i_\tau=(p^i_{x,\tau},p^i_{y,\tau})$. Nearest neighbor search and reconstruction of the objective functions are performed at each iteration of optimization. As a result, the vehicle that already takes the lead will set its own reference waypoints further ahead, while the vehicle that falls behind does the opposite. Together with the safety constraints, they are iteratively separated; and as a result, the passing sequence is determined.

The nearest neighbor search for the closest waypoint is also implemented using the k-d tree for improved efficiency. Constant reference velocity $v_{ref}$ is used for all vehicles at all time stamps, and no restriction on heading angles is imposed. Moreover, covariance matrix $Q^i_\tau$ is defined such that only lateral deviation from the reference path is considered, while longitudinal deviation from the reference waypoint along the reference path is ignored.

\subsection{Optimization Problem Reformulation}

To handle the nonlinear kinematics, we perform the first-order Taylor expansion of (\ref{shortDynamics}) to obtain
\begin{equation}
\label{linearKinematics}
    \delta x^i_{\tau+1}=A^i_\tau\delta x^i_\tau+B^i_\tau \delta u^i_\tau.
\end{equation}
Input constraints (\ref{inputConstraints}) are reformulated as
\begin{equation}
\label{inputConstraints}
    a_\textup{min}-a^i_\tau \leq \delta a^i_\tau\leq a_\textup{max}-a^i_\tau, 
    \delta_\textup{min}-\delta^i_\tau \leq\delta\delta^i_\tau\leq\delta_\textup{max}-\delta^i_\tau.
\end{equation}

We tile up (\ref{collisionAvoidanceAll}), (\ref{boundaryConstraints}), and (\ref{inputConstraints}) and rewrite them into a compact matrix form, which yields
\begin{equation}
\label{Gfunction}
    \sum\limits_{i=1}^N J^i\delta X^i+l\geq 0.
\end{equation}

We denote the indicator function over a given set $\mathcal{S}$ as $\mathcal{I}_\mathcal{S}(x)$. With this notation, we gather (\ref{Gfunction}), objective functions, and linearized kinematics to obtain the following quadratic optimization problem
\begin{equation}
\label{newProb}
    \min\limits_{\{\delta X^i\}}\sum\limits_{i=1}^N\left(C^i(\delta X^i)+\mathcal{I}_{\mathcal{K}^i}(\delta X^i)\right)+\mathcal{I}_{R^n_+}\left(\sum\limits_{i=1}^NJ^i\delta X^i+l\right),
\end{equation}
where $\mathcal{K}^i$ is the feasible set for linearized kinematics of vehicle $i$ and ${R^n_+}$ is the nonnegative orthant with $n$ being the length of $l$.

\subsection{Parallel Optimization}
Dual consensus ADMM is adopted to perform parallel optimization. 
We define the proximal operator as $\textup{Prox}^\rho_f(x)=\arg\min\limits_{y}\{f(y)+\frac{\rho}{2}||x-y||^2\},$ 
and Algorithm 1~\cite{banjac2019decentralized} can be used to solve the following convex optimization problem:
\begin{equation}
\min\limits_{\{x^i\}}\ \sum\limits_{i=1}^N f^i(x^i)+g\left(\sum\limits_{i=1}^NJ^ix^i\right).
\end{equation}

\begin{algorithm}[t]
\caption{Dual Consensus ADMM~\cite{banjac2019decentralized}}\label{alg:alg1}
\begin{algorithmic}[1]
\State \textbf{choose} $\sigma,\rho >0$
\State \textbf{initialize} for all $i \in \mathcal{N}$: $p^{i,0}=y^{i,0}=z^{i,0}=s^{i,0}=0$
\State \textbf{repeat}: for all $i \in \mathcal{N}$
\State \hspace{0.25cm} Broadcast $y^{i,k}$ to all other vehicles
\State \hspace{0.25cm} $p^{i,k+1}\leftarrow p^{i,k}+\rho\sum_{j\neq i}(y^{i,k}-y^{j,k})$
\State \hspace{0.25cm} $s^{i,k+1}\leftarrow s^{i,k}+\sigma(y^{i,k}-z^{i,k})$
\State \hspace{0.25cm} $r^{i,k+1}\leftarrow \rho\sum_{j\neq i}(y^{i,k}+y^{j,k})+\sigma z^{i,k}-p^{i,k+1}-s^{i,k+1}$
\State \hspace{0.25cm} $x^{i,k+1}\leftarrow \underset{{x^i}}{\arg\min}\{f^i(x^i)+\eta||J^i x^i+r^{i,k+1}||^2\}$
\State \hspace{0.25cm} $y^{i,k+1}\leftarrow 2\eta(J^i x^{i,k+1}+r^{i,k+1})$
\State \hspace{0.25cm} $z^{i,k+1,*}\leftarrow \textup{Prox}^{\frac{1}{N\sigma}}_g(N(s^{i,k+1}+\sigma y^{i,k+1}))$
\State \hspace{0.25cm} $z^{i,k+1}\leftarrow \frac{s^{i,k+1}}{\sigma}+y^{i,k+1}-\frac{1}{N\sigma}z^{i,k+1,*}$
\State \textbf{until} termination criterion is satisfied
\end{algorithmic}
\label{alg1}
\end{algorithm}

Note that ${p,s,r,y,z}$ are column vectors of the same size as $l$, and $\eta=1/(2(\sigma+2\rho(N-1)))$. We apply dual consensus ADMM to problem (\ref{newProb}) with the following substitution: $x^i\leftarrow \delta X^i$, $f^i(\boldsymbol{\cdot})\leftarrow C^i(\boldsymbol{\cdot})+\mathcal{I}_{\mathcal{K}^i}(\boldsymbol{\cdot})$, and $g(\boldsymbol{\cdot})\leftarrow\mathcal{I}_{R^n_+}(\boldsymbol{\cdot}+l)$. Detailed discussions specific to our problem are provided as follows.

\subsubsection{Dual Update}
In the dual consensus ADMM algorithm, variables $\{y^i\}$ and $\{z^i\}$ are dual variables to the original optimization problem. The updates of $\{y^i\}$ and $\{z^i\}$ are performed by Steps 5-7 and 9-11 of Algorithm 1. In particular, Step 10 is performed by
\begin{equation}
\label{zStarUpdate}
    z^{i,k+1,*}=\textup{max}\{N(s^{i,k+1}+\sigma y^{i,k+1}),-l+\epsilon\},
\end{equation}
{which is derived from the definition of the proximal operator.} $\epsilon$ is a small non-negative number to push $z$ away from the boundary into the feasible region so as to guarantee strict satisfaction of constraints. $\textup{max}\{x,y\}$ denotes the element-wise maximization between two vectors $x$ and $y$.

\subsubsection{Primal Update}
The primal update is performed by Step 8 of Algorithm 1 to search for the optimal states and control inputs for vehicle $i$. We perform expansion of the square term and obtain
\begin{equation}
\label{deltaX}
\begin{aligned}
    \delta X^{i,k+1} &= \underset{{\delta X^i}}{\arg\min}\{C^i(\delta X^i)+\mathcal{I}_{\mathcal{K}^i}(\delta X^i)\\
    &+\eta(\delta X^{i\top}J^{i\top}J^i\delta X^i+2r^{i,k+1\top}J^i\delta X^i)\}.
\end{aligned}
\end{equation}

We perform further expansion of the third term, and the resulting collision avoidance term is
\begin{equation}
\begin{aligned}
    C^i_c(\delta X^i)&=\eta\sum_{\tau=0}^{T}\delta x^{i\top}_\tau P^i_{c,\tau}\delta x^i_\tau+2\Gamma^i_{c,\tau}\delta x^i_\tau,\\    P^i_{c,\tau}&=\sum_{j,\beta,\gamma}J^{i\top}_{\beta,\tau}\vec{n}^{ij}_{\beta\gamma,\tau}\vec{n}^{ij\top}_{\beta\gamma,\tau}J^{i}_{\beta,\tau},\\    \Gamma^i_{c,\tau}&=\sum_{j,\beta,\gamma}r^{i,k+1}_{c,\beta\gamma,j,\tau}\vec{n}^{ij\top}_{\beta\gamma,\tau}J^{i}_{\beta,\tau},\forall j\neq i,\beta,\gamma\in\{f,r\}.
\end{aligned}
\end{equation}
The resulting road boundary term is
\begin{equation}
\begin{aligned}
    C^i_b(\delta X^i)&=\eta\sum_{\tau=0}^{T}\delta x^{i\top}_\tau P^i_{b,\tau}\delta x^i_\tau+2\Gamma^i_{b,\tau}\delta x^i_\tau,\\
    P^i_{b,\tau}&=\sum_{\beta\in\{f,r\}}J^{i\top}_{\beta,\tau}J^{i}_{\beta,\tau}, \Gamma^i_{b,\tau}=\sum_{\beta\in\{f,r\}}r^{i,k+1}_{b,\beta,\tau}\vec{n}^{i\top}_{\beta,\tau}J^{i}_{\beta,\tau}.
\end{aligned}
\end{equation}
The resulting control input term is
\begin{equation}
    C_{ci}^i(\delta X^i)=\eta\sum_{\tau=0}^{T-1}\delta u^{i\top}_\tau\delta u^{i}_\tau+2r^{i,k+1\top}_{ci,\tau}\delta u^i_\tau.
\end{equation}
In the above equations, $r^{i,k+1}_{c,\beta\gamma,j,\tau}$ is the element of $r^{i,k+1}$ that locates at the same row as the collision avoidance constraint between $\beta$ circle of vehicle $i$ and $\gamma$ circle of vehicle $j$ at timestamp $\tau$. $r^{i,k+1}_{b,\beta,\tau}$ and $r^{i,k+1}_{ci,\tau}$ are defined similarly for road boundary constraints and control input constraints, respectively. With the above terms, (\ref{deltaX}) can be rewritten as
\begin{equation}
\label{LQR}
\begin{aligned}
    \min\limits_{\delta X^i}\ &C^i(\delta X^i)+C^i_c(\delta X^i)+C^i_b(\delta X^i)+C^i_{ci}(\delta X^i)\\
    \textup{s.t.}\ &\delta x^i_{\tau+1}=A^i_\tau\delta x^i_\tau + B^i_\tau\delta u^i_\tau.
\end{aligned}
\end{equation}

\begin{algorithm}[t]
\caption{Parallel Optimization Algorithm for Cooperative Driving with Hard Safety Constraints}\label{alg:alg1}
\begin{algorithmic}[1]
\State \textbf{for} $i\leftarrow 1$ \textbf{to} $N$ \textbf{in parallel}:
\State \hspace{0.25cm} \textbf{initialize} $\{x^i_\tau,u^i_\tau\}^T_{\tau=0},\{p^{i,0},y^{i,0},z^{i,0},s^{i,0}\},\sigma,\rho$
\State \hspace{0.25cm} \textbf{repeat}:
\State \hspace{0.50cm} Send $\{x^i_\tau\}_{\tau=1}^T$ to $j\in\mathcal{N}-\{i\}$
\State \hspace{0.50cm} Compute $\{J^i_{f,\tau},J^i_{r,\tau},p^i_{f,\tau},p^i_{r,\tau}\}^{T}_{\tau=0}$, $\{A^i_\tau,B^i_\tau\}^{T-1}_{\tau=0}$
\State \hspace{0.50cm} \textbf{for} $\tau\leftarrow 0$ \textbf{to} $T$:
\State \hspace{0.75cm} \textbf{for} $j$ \textbf{in} $\mathcal{N}-\{i\}$:
\State \hspace{1.0cm} Compute $\{p^j_{f,\tau},p^j_{r,\tau}\}$
\State \hspace{1.0cm} \textbf{for} $\beta,\gamma$ \textbf{in} $\{f,r\}$:
\State \hspace{1.25cm} $\vec{n}^{ij}_{\beta\gamma,\tau}\leftarrow(p^i_{\beta,\tau}-p^j_{\gamma,\tau})/||p^i_{\beta,\tau}-p^j_{\gamma,\tau}||_2$
\State \hspace{1.0cm} \textbf{end}
\State \hspace{0.75cm} \textbf{end}
\State \hspace{0.75cm} \textbf{for} $\beta$ \textbf{in} $\{f,r\}$:
\State \hspace{1.0cm} $p^i_{\beta b,\tau}\leftarrow NearestNeighbor(\mathcal{M},p^i_{\beta,\tau})$
\State \hspace{1.0cm} $\vec{n}^i_{\beta,\tau}\leftarrow(p^i_{\beta,\tau}-p^i_{\beta b,\tau})/||p^i_{\beta,\tau}-p^i_{\beta b,\tau}||_2$
\State \hspace{0.75cm} \textbf{end}
\State \hspace{0.75cm} $p^i_{\tau,ref}\leftarrow NearestNeighbor(\{p^i_{ref}\},p^i_\tau)$
\State \hspace{0.75cm} Compute $Q^i_\tau$
\State \hspace{0.5cm} \textbf{end}
\State \hspace{0.5cm} Compute $\{P^i_{c,\tau},P^i_{b,\tau}\}_{\tau=0}^{T},J^i,l$
\State \hspace{0.5cm} $p^{i,0}\leftarrow 0,s^{i,0}\leftarrow 0,y^{i,0}\leftarrow y^{prev},z^{i,0}\leftarrow z^{prev}$
\State \hspace{0.5cm} \textbf{for} $k\leftarrow 0$ \textbf{to} $k_\textup{max}$:
\State \hspace{0.75cm} Steps 4-7 of \textbf{Algorithm 1}
\State \hspace{0.75cm} Compute $\{\Gamma^i_{c,\tau},\Gamma^i_{b,\tau}\}^T_{\tau=0}$
\State \hspace{0.75cm} $\delta X^{i,k+1}\leftarrow$ Problem (\ref{LQR})
\State \hspace{0.75cm} $y^{i,k+1}\leftarrow 2\eta(J^i \delta X^{i,k+1}+r^{i,k+1})$
\State \hspace{0.75cm} $z^{i,k+1}\leftarrow\frac{s^{i,k+1}}{\sigma}+y^{i,k+1}$
\Statex \hspace{1.25cm} $-\frac{1}{N\sigma}\textup{max}\{N(s^{i,k+1}+\sigma y^{i,k+1}),-l+\epsilon\}$
\State \hspace{0.5cm} \textbf{end}
\State \hspace{0.5cm} $y^{prev}\leftarrow y^{i,k+1}, z^{prev}\leftarrow z^{i,k+1}$
\State \hspace{0.5cm} Update $\{x^i_\tau,u^i_\tau\}^T_{\tau=0}$
\State \hspace{0.25cm} \textbf{until} termination criterion is satisfied
\State \textbf{end}
\end{algorithmic}
\label{alg2}
\end{algorithm}

Obviously, (\ref{LQR}) is a standard LQR problem, which can be solved by dynamic programming effectively. Note that (\ref{LQR}) only involves state variables and control inputs confined to one vehicle, and therefore (\ref{LQR}) corresponding to different vehicles can be solved in a parallel manner.

With the above discussions, we propose Algorithm 2 for solving problem (\ref{OCP}). Convex reformulation of problem (\ref{OCP}) and loops of dual consensus ADMM are performed in an alternating manner until the termination criterion is met.
The update of trajectory is performed following the same method in~\cite{huang2023decentralized}. Given the convexity of (\ref{newProb}), ADMM is guaranteed to converge, and thus consensus will be established after a sufficiently large number of iterations. In practice, we let the algorithm terminate when the trajectories are completely collision-free and the change in the overall cost between two consecutive iterations is smaller than a given criterion $\zeta$.

\begin{table}[t]
\centering
\caption{Parameters used in experiments}
\begin{tabular}{p{0.7cm}p{1.28cm}p{0.7cm}p{1.0cm}p{0.7cm}p{1.0cm}}
\hline
Param.               & Value                            & Param.              & Value 
& Param.              & Value \\ \hline

$b$                      & 3.0\,m                         & $v_{ref}$              & 10\,m/s
&$\tau_s $                & 0.1\,s \\

$a_\textup{max} $       & 8.0\,$\textup{m}/\textup{s}^2$ &  $\sigma$                 & 0.2 & $\rho$                   & 0.02  \\

$a_\textup{min}$         & -12.0\,$\textup{m}/\textup{s}^2$ & $\delta_\textup{max}$   & 0.62\,rad  
&  $\epsilon$             & 0.3 \\                     

$k_\textup{max}$         & 2                      &$\delta_\textup{min}$    & -0.62\,rad                       & $\zeta $               & 1.0   \\

$d_\textup{safe}$        & 2.62\,m                          & $T$                    & 75  
&$d_f$                   & 2.79\,m   \\   

$d_r$                    & -0.05\,m  &                         &                        &       \\ \hline
\end{tabular}
\end{table}

\section{Experimental Results}

To verify the proposed method, we perform simulations on the roundabout area in Town03, CARLA simulator~\cite{dosovitskiy2017carla}. Altogether 16 CAVs are involved in this scenario. For all CAVs, reference paths are pre-generated by the global navigation module in CARLA.
The reference paths are identical for all methods and fixed throughout the entire planning process. Key parameter settings are shown in Table I, where $d_f$ and $d_r$ are the biases between the rear axle center and centers of $FC$ and $RC$. Initial velocities are 10\,m/s for all CAVs.


\subsection{Computation Efficiency}

 To demonstrate the improved performance in terms of computation efficiency, we compare the computation time of our method with the general IPOPT solver provided by CasADi~\cite{andersson2019casadi}. Both our method and the IPOPT solver are implemented on a server with $2\times$ Intel(R) Xeon(R) Gold 6348 CPU @ 2.60 GHz. For our method, both single-process and multi-process implementations are provided. For the latter one, the number of processes is set to be $N$.

 One difficulty in solving (\ref{OCP}) with IPOPT solver is that the reference points need to be defined during the formulation of the problem, which are unknown at the beginning. We simplify (\ref{OCP}) for the IPOPT solver by first solving it using our method, and then formulating the problem with the optimal reference points obtained. Meanwhile, we examine two different solving schemes. The one-stage scheme directly optimizes (\ref{OCP}) with full constraints, while the two-stage scheme first optimizes (\ref{OCP}) with collision avoidance constraints released, and then uses the obtained results as initialization for optimization of the whole problem.


Table II compares the computational efficiency between all four optimization settings. We use the computation time per timestamp as a measure. It can be seen that the multi-process implementation of our method is the fastest of all, which is at least two orders of magnitude faster than the IPOPT solver. This result clearly demonstrates the superiority in computational efficiency of our method for solving cooperative trajectory planning problems. Meanwhile, the multi-process implementation of our method is $4.9\times$, $5.7\times$, and $5.5\times$ as fast as the single process implementation for $N=8, 12, \textup{and} \, 16$, respectively. These results demonstrate the effectiveness of parallel optimization.

 
\begin{table}[t]
\centering
\caption{Computation time per time stamp with different methods}
\begin{tabular}{lcccc}
\hline
 & \multicolumn{2}{c}{Our Method} & \multicolumn{2}{c}{IPOPT solver} \\
 & Single-process         & Multi-process         & Two-stage         & One-stage         \\ \hline
$N$=8 & 0.00772\,s         & \textbf{0.00156\,s}         & 0.109\,s         & 0.153\,s \\ 
$N$=12 & 0.0235\,s         & \textbf{0.00411\,s}         & 0.235\,s         & 4.37\,s \\ 
$N$=16 & 0.0480\,s         & \textbf{0.00876\,s}         & 0.506\,s         & 7.09\,s \\ \hline
\end{tabular}
\end{table}

\subsection{Traffic Efficiency in Roundabout Area}

\begin{figure}[t]


\centering
\subfigure[$\tau=1.0\,\textup{s}$ ours]{\includegraphics[scale=0.105]{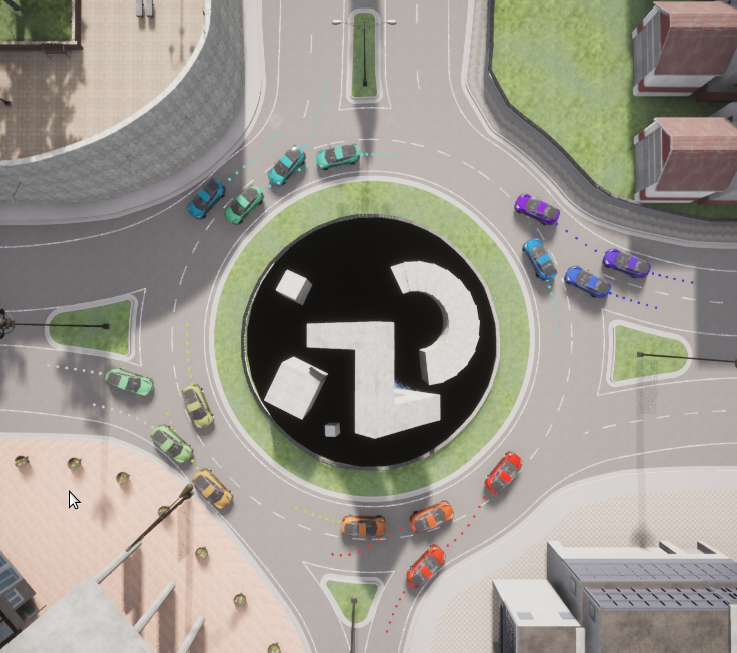}}
\subfigure[$\tau=3.0\,\textup{s}$ ours]{\includegraphics[scale=0.105]{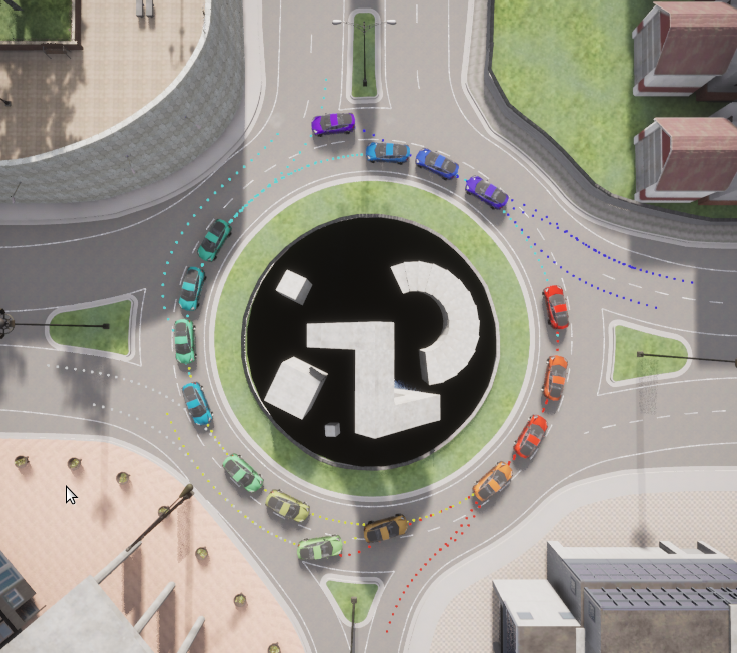}}
\subfigure[$\tau=5.0\,\textup{s}$ ours]{\includegraphics[scale=0.105]{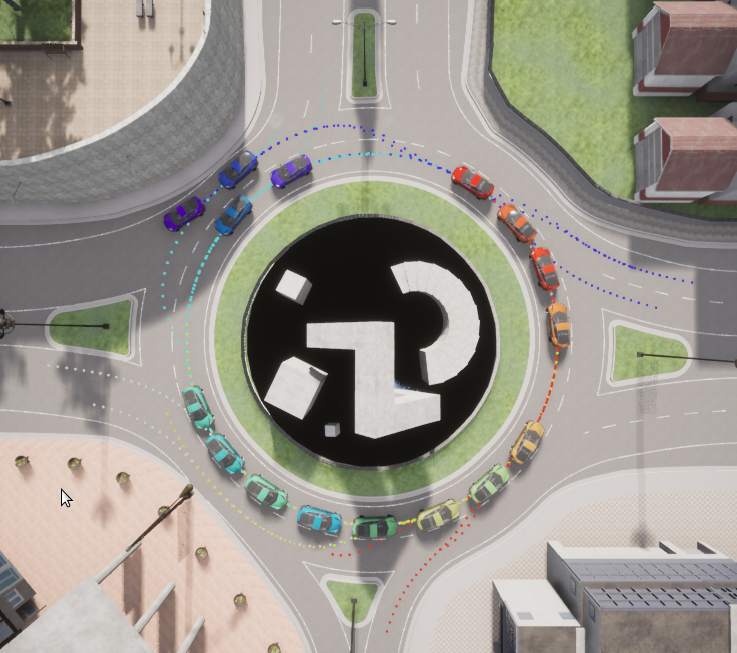}}


\subfigure[$\tau=1.0\,\textup{s}$ baseline]{\includegraphics[scale=0.105]{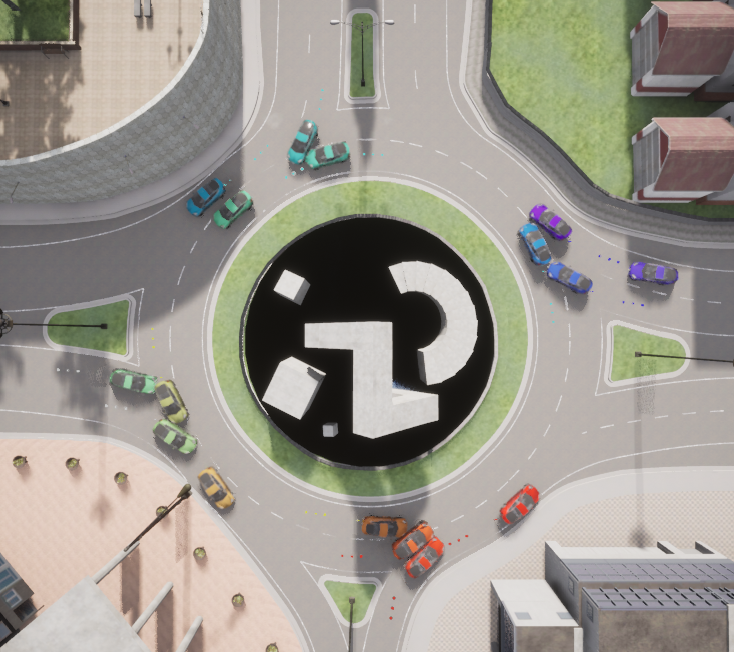}}
\subfigure[$\tau=3.0\,\textup{s}$ baseline]{\includegraphics[scale=0.105]{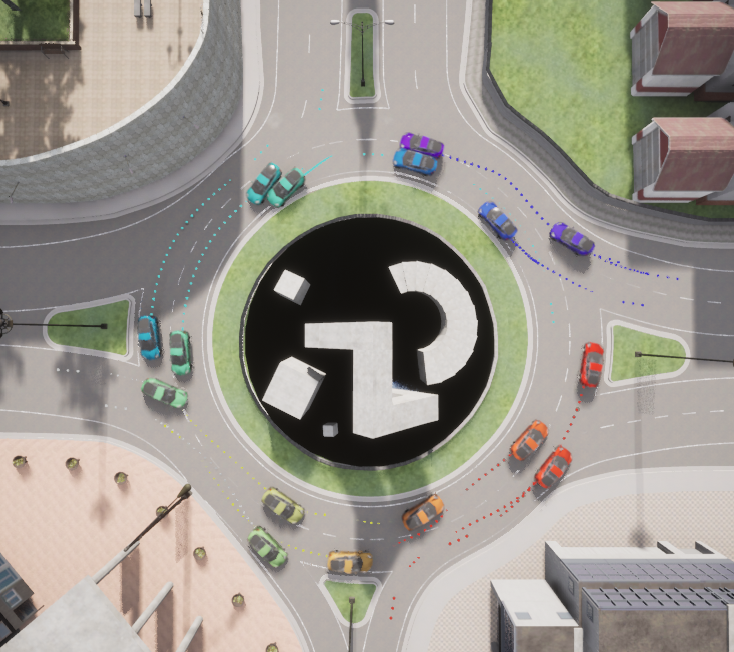}}
\subfigure[$\tau=5.0\,\textup{s}$ baseline]{\includegraphics[scale=0.105]{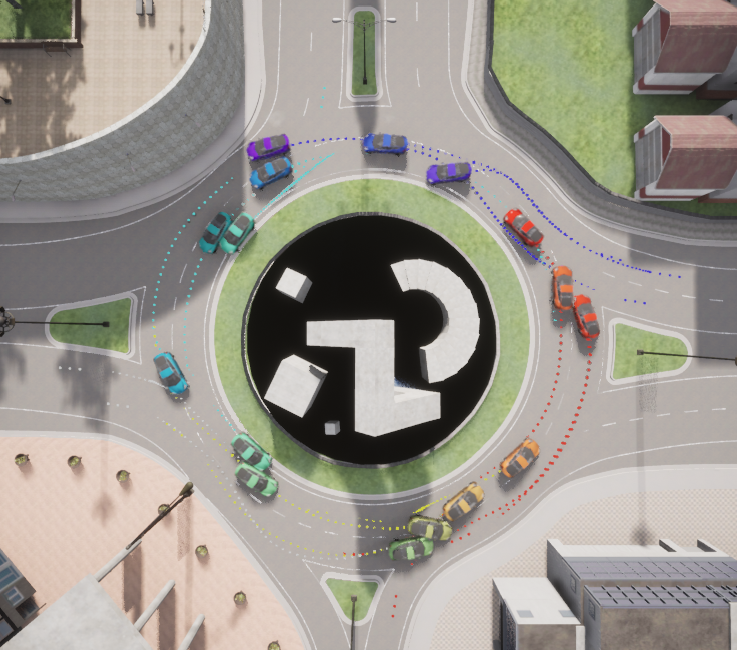}}

\caption{Simulation results of our method and baseline at time equal to 1.0\,s, 3.0\,s, and 5.0\,s, respectively.}
\label{fig:simulations1}
\end{figure}





To demonstrate the overall passing efficiency, we compare our method with the baseline proposed by~\cite{dosovitskiy2017carla}, which tracks the reference path using a low-level controller and brakes to maintain safe distance with surrounding vehicles. Qualitative results in Fig. \ref{fig:simulations1} show the positions of vehicles and past trajectories at selected timestamps for our method and the baseline, respectively. Following the given reference paths, smooth and efficient merging into the inner lane is observed in the trajectories generated by our method, which reduces the overall travel distances. Meanwhile, collisions between vehicles are avoided. On the contrary, the baseline exhibits low efficiency in passing and fails to avoid collisions.

Quantitatively, we compare the resulting average velocities of vehicles by our method and the baseline. The vehicles are naturally divided into 4 groups according to the entrances they start from. For each group of vehicles, we compute their average velocities over the entire horizon. The results are shown in {Table III}. It is clear that for all cases, all 4 groups of vehicles run at higher average velocities with our method than that with the baseline. The average velocities of vehicles attained by our method are close to the reference velocity $v_\textup{ref}=10\,\textup{m}/\textup{s}$, while the slowdown is observed in the baseline. These results show that our method yields higher overall passing efficiency for this crowded roundabout area.
\begin{table}[t]
\centering
\caption{Average velocity of vehicles with different methods}
\begin{tabular}{p{0.5cm}p{1.0cm}cccc}
\hline
&        &Group 1   & Group 2   & Group 3   & Group 4 \\ \hline

\multirow{2}{*}{$N$=8}& Baseline    
& 7.35\,$\textup{m}/\textup{s}$     & 7.81\,$\textup{m}/\textup{s}$     
& 7.80\,$\textup{m}/\textup{s}$     & 7.65\,$\textup{m}/\textup{s}$ \\
& Proposed                                      
& \textbf{10.0}\,$\textup{m}/\textup{s}$    & \textbf{9.14}\,$\textup{m}/\textup{s}$     
& \textbf{9.85}\,$\textup{m}/\textup{s}$    & \textbf{9.33}\,$\textup{m}/\textup{s}$ \\ \hline

\multirow{2}{*}{$N$=12}& Baseline
& 7.49\,$\textup{m}/\textup{s}$     & 7.74\,$\textup{m}/\textup{s}$     
& 6.92\,$\textup{m}/\textup{s}$     & 7.53\,$\textup{m}/\textup{s}$ \\
& Proposed                                       
& \textbf{10.0}\,$\textup{m}/\textup{s}$    & \textbf{9.36}\,$\textup{m}/\textup{s}$     
& \textbf{9.27}\,$\textup{m}/\textup{s}$    & \textbf{9.40}\,$\textup{m}/\textup{s}$ \\ \hline

\multirow{2}{*}{$N$=16}& Baseline
& 7.25\,$\textup{m}/\textup{s}$     & 5.51\,$\textup{m}/\textup{s}$     
& 6.37\,$\textup{m}/\textup{s}$     & 7.21\,$\textup{m}/\textup{s}$ \\
& Proposed                                      
& \textbf{9.86}\,$\textup{m}/\textup{s}$    & \textbf{9.39}\,$\textup{m}/\textup{s}$     
& \textbf{9.14}\,$\textup{m}/\textup{s}$    & \textbf{9.08}\,$\textup{m}/\textup{s}$ \\ \hline

\end{tabular}
\end{table}

\section{Conclusions}

In this paper, we investigate the cooperative autonomous driving problem of CAVs in urban traffic scenarios. We formulate the problem as an optimal control problem with safety conditions enforced as hard constraints, together with other constraints including vehicle dynamics and control limitations. Then, a parallel optimization framework is presented to solve the specific problem efficiently. We compare our method with different baselines to demonstrate the effectiveness in terms of average travel efficiency and to show the superiority in computational efficiency. 
Possible future works include the incorporation of a high-level decision-making strategy into the optimization framework to further enhance traffic efficiency.
We also consider extending the optimization method to a mixed traffic flow that contains both CAVs and human-driving vehicles.





\bibliographystyle{IEEEtran}
\bibliography{refs}

\begin{thebibliography}{10}
\providecommand{\url}[1]{#1}
\csname url@samestyle\endcsname
\providecommand{\newblock}{\relax}
\providecommand{\bibinfo}[2]{#2}
\providecommand{\BIBentrySTDinterwordspacing}{\spaceskip=0pt\relax}
\providecommand{\BIBentryALTinterwordstretchfactor}{4}
\providecommand{\BIBentryALTinterwordspacing}{\spaceskip=\fontdimen2\font plus
\BIBentryALTinterwordstretchfactor\fontdimen3\font minus
  \fontdimen4\font\relax}
\providecommand{\BIBforeignlanguage}[2]{{%
\expandafter\ifx\csname l@#1\endcsname\relax
\typeout{** WARNING: IEEEtran.bst: No hyphenation pattern has been}%
\typeout{** loaded for the language `#1'. Using the pattern for}%
\typeout{** the default language instead.}%
\else
\language=\csname l@#1\endcsname
\fi
#2}}
\providecommand{\BIBdecl}{\relax}
\BIBdecl

\bibitem{deng2019cooperative}
R.~Deng, B.~Di, and L.~Song, ``Cooperative collision avoidance for overtaking
  maneuvers in cellular {V2X}-based autonomous driving,'' \emph{IEEE
  Transactions on Vehicular Technology}, vol.~68, no.~5, pp. 4434--4446, 2019.

\bibitem{liu2017distributed}
C.~Liu, C.-W. Lin, S.~Shiraishi, and M.~Tomizuka, ``Distributed conflict
  resolution for connected autonomous vehicles,'' \emph{IEEE Transactions on
  Intelligent Vehicles}, vol.~3, no.~1, pp. 18--29, 2017.

\bibitem{hang2021cooperative}
P.~Hang, C.~Lv, C.~Huang, Y.~Xing, and Z.~Hu, ``Cooperative decision making of
  connected automated vehicles at multi-lane merging zone: A coalitional game
  approach,'' \emph{IEEE Transactions on Intelligent Transportation Systems},
  vol.~23, no.~4, pp. 3829--3841, 2021.

\bibitem{rios2016automated}
J.~Rios-Torres and A.~A. Malikopoulos, ``Automated and cooperative vehicle
  merging at highway on-ramps,'' \emph{IEEE Transactions on Intelligent
  Transportation Systems}, vol.~18, no.~4, pp. 780--789, 2016.

\bibitem{ding2019rule}
J.~Ding, L.~Li, H.~Peng, and Y.~Zhang, ``A rule-based cooperative merging
  strategy for connected and automated vehicles,'' \emph{IEEE Transactions on
  Intelligent Transportation Systems}, vol.~21, no.~8, pp. 3436--3446, 2019.

\bibitem{xu2019grouping}
H.~Xu, S.~Feng, Y.~Zhang, and L.~Li, ``A grouping-based cooperative driving
  strategy for {CAV}s merging problems,'' \emph{IEEE Transactions on Vehicular
  Technology}, vol.~68, no.~6, pp. 6125--6136, 2019.

\bibitem{mirheli2019consensus}
A.~Mirheli, M.~Tajalli, L.~Hajibabai, and A.~Hajbabaie, ``A consensus-based
  distributed trajectory control in a signal-free intersection,''
  \emph{Transportation Research Part C: Emerging Technologies}, vol. 100, pp.
  161--176, 2019.

\bibitem{fayazi2018mixed}
S.~A. Fayazi and A.~Vahidi, ``Mixed-integer linear programming for optimal
  scheduling of autonomous vehicle intersection crossing,'' \emph{IEEE
  Transactions on Intelligent Vehicles}, vol.~3, no.~3, pp. 287--299, 2018.

\bibitem{chen2019model}
J.~Chen, B.~Yuan, and M.~Tomizuka, ``Model-free deep reinforcement learning for
  urban autonomous driving,'' in \emph{2019 IEEE Intelligent Transportation
  Systems Conference (ITSC)}.\hskip 1em plus 0.5em minus 0.4em\relax IEEE,
  2019, pp. 2765--2771.

\bibitem{guan2020centralized}
Y.~Guan, Y.~Ren, S.~E. Li, Q.~Sun, L.~Luo, and K.~Li, ``Centralized cooperation
  for connected and automated vehicles at intersections by proximal policy
  optimization,'' \emph{IEEE Transactions on Vehicular Technology}, vol.~69,
  no.~11, pp. 12\,597--12\,608, 2020.

\bibitem{toghi2021cooperative}
B.~Toghi, R.~Valiente, D.~Sadigh, R.~Pedarsani, and Y.~P. Fallah, ``Cooperative
  autonomous vehicles that sympathize with human drivers,'' in \emph{2021
  IEEE/RSJ International Conference on Intelligent Robots and Systems
  (IROS)}.\hskip 1em plus 0.5em minus 0.4em\relax IEEE, 2021, pp. 4517--4524.

\bibitem{pan2022convex}
X.~Pan, B.~Chen, S.~Timotheou, and S.~A. Evangelou, ``A convex optimal control
  framework for autonomous vehicle intersection crossing,'' \emph{arXiv
  preprint arXiv:2203.16870}, 2022.

\bibitem{xu2019tree}
H.~Xu, Y.~Zhang, L.~Li, and W.~Li, ``Cooperative driving at unsignalized
  intersections using tree search,'' \emph{IEEE Transactions on Intelligent
  Transportation Systems}, vol.~21, no.~11, pp. 4563--4571, 2019.

\bibitem{kessler2019cooperative}
T.~Kessler and A.~Knoll, ``Cooperative multi-vehicle behavior coordination for
  autonomous driving,'' in \emph{2019 IEEE Intelligent Vehicles Symposium
  (IV)}.\hskip 1em plus 0.5em minus 0.4em\relax IEEE, 2019, pp. 1953--1960.

\bibitem{esterle2020optimal}
K.~Esterle, T.~Kessler, and A.~Knoll, ``Optimal behavior planning for
  autonomous driving: A generic mixed-integer formulation,'' in \emph{2020 IEEE
  Intelligent Vehicles Symposium (IV)}.\hskip 1em plus 0.5em minus 0.4em\relax
  IEEE, 2020, pp. 1914--1921.

\bibitem{burger2018cooperative}
C.~Burger and M.~Lauer, ``Cooperative multiple vehicle trajectory planning
  using {MIQP},'' in \emph{2018 21st International Conference on Intelligent
  Transportation Systems (ITSC)}.\hskip 1em plus 0.5em minus 0.4em\relax IEEE,
  2018, pp. 602--607.

\bibitem{boyd2011distributed}
S.~Boyd, N.~Parikh, E.~Chu, B.~Peleato, and J.~a. Eckstein, ``Distributed
  optimization and statistical learning via the alternating direction method of
  multipliers,'' \emph{Foundations and Trends{\textregistered} in Machine
  Learning}, vol.~3, no.~1, pp. 1--122, 2011.

\bibitem{rey2018fully}
F.~Rey, Z.~Pan, A.~Hauswirth, and J.~Lygeros, ``Fully decentralized {ADMM} for
  coordination and collision avoidance,'' in \emph{2018 European Control
  Conference (ECC)}.\hskip 1em plus 0.5em minus 0.4em\relax IEEE, 2018, pp.
  825--830.

\bibitem{wang2018parallel}
Z.~Wang, Y.~Zheng, S.~E. Li, K.~You, and K.~Li, ``Parallel optimal control for
  cooperative automation of large-scale connected vehicles via {ADMM},'' in
  \emph{2018 21st International Conference on Intelligent Transportation
  Systems (ITSC)}.\hskip 1em plus 0.5em minus 0.4em\relax IEEE, 2018, pp.
  1633--1639.

\bibitem{zhang2021semi}
X.~Zhang, Z.~Cheng, J.~Ma, S.~Huang, F.~L. Lewis, and T.~H. Lee,
  ``Semi-definite relaxation-based {ADMM} for cooperative planning and control
  of connected autonomous vehicles,'' \emph{IEEE Transactions on Intelligent
  Transportation Systems}, vol.~23, no.~7, pp. 9240--9251, 2021.

\bibitem{saravanos2022distributed}
A.~D. Saravanos, Y.~Aoyama, H.~Zhu, and E.~A. Theodorou, ``Distributed
  differential dynamic programming architectures for large-scale multi-agent
  control,'' \emph{arXiv preprint arXiv:2207.13255}, 2022.

\bibitem{banjac2019decentralized}
G.~Banjac, F.~Rey, P.~Goulart, and J.~Lygeros, ``Decentralized resource
  allocation via dual consensus {ADMM},'' in \emph{2019 American Control
  Conference (ACC)}.\hskip 1em plus 0.5em minus 0.4em\relax IEEE, 2019, pp.
  2789--2794.

\bibitem{huang2023decentralized}
Z.~Huang, S.~Shen, and J.~Ma, ``Decentralized i{LQR} for cooperative trajectory
  planning of connected autonomous vehicles via dual consensus {ADMM},''
  \emph{arXiv preprint arXiv:2301.04386}, 2023.

\bibitem{tassa2014control}
Y.~Tassa, N.~Mansard, and E.~Todorov, ``Control-limited differential dynamic
  programming,'' in \emph{2014 IEEE International Conference on Robotics and
  Automation (ICRA)}.\hskip 1em plus 0.5em minus 0.4em\relax IEEE, 2014, pp.
  1168--1175.

\bibitem{dosovitskiy2017carla}
A.~Dosovitskiy, G.~Ros, F.~Codevilla, A.~Lopez, and V.~Koltun, ``{CARLA}: An
  open urban driving simulator,'' in \emph{Conference on Robot Learning}.\hskip
  1em plus 0.5em minus 0.4em\relax PMLR, 2017, pp. 1--16.

\bibitem{andersson2019casadi}
J.~A. Andersson, J.~Gillis, G.~Horn, J.~B. Rawlings, and M.~Diehl,
  ``{C}as{AD}i: a software framework for nonlinear optimization and optimal
  control,'' \emph{Mathematical Programming Computation}, vol.~11, pp. 1--36,
  2019.

\end{thebibliography}

\end{document}